\documentclass[10pt,twocolumn,letterpaper]{article}

\usepackage{cvpr}
\usepackage{times}
\usepackage{epsfig}
\usepackage{graphicx}
\usepackage{subcaption}
\usepackage{amsmath}
\usepackage{amssymb}
\usepackage{pifont}
\newcommand{\cmark}{\ding{51}}%
\newcommand{\xmark}{\ding{55}}%
\usepackage{mathtools}

\usepackage{bm}
\usepackage{bbm}
\usepackage{multirow}
\usepackage{multicol}
\usepackage{gensymb}
\usepackage{graphics}

\usepackage[dvipsnames]{xcolor}
\definecolor{mygray}{gray}{15}
\usepackage[super]{nth}
\usepackage[normalem]{ulem}

\usepackage[pagebackref=true,breaklinks=true,letterpaper=true,colorlinks,bookmarks=false]{hyperref}

\cvprfinalcopy 

\def\cvprPaperID{16} 

\ifcvprfinal\fi
\begin{document}

\title{\nth{1} Place Solution for\\ Waymo Open Dataset Challenge - 3D Detection and Domain Adaptation}

\author{
Zhuangzhuang Ding$^{*}$ \quad Yihan Hu$^{*}$ \quad Runzhou Ge$^{*}$ \\
Li Huang \quad Sijia Chen \quad Yu Wang \quad Jie Liao \\
Horizon Robotics\\
{\tt\small \{dinghouzx, yihan.hu96, runzhouge, yuwangrpi\}@gmail.com} 
}

\twocolumn[{%
\vspace{-1em}
\maketitle
\vspace{-1em}

\begin{center}
    \centering 
    
    \vspace{-0.3in}
    \includegraphics[width=0.95\textwidth]{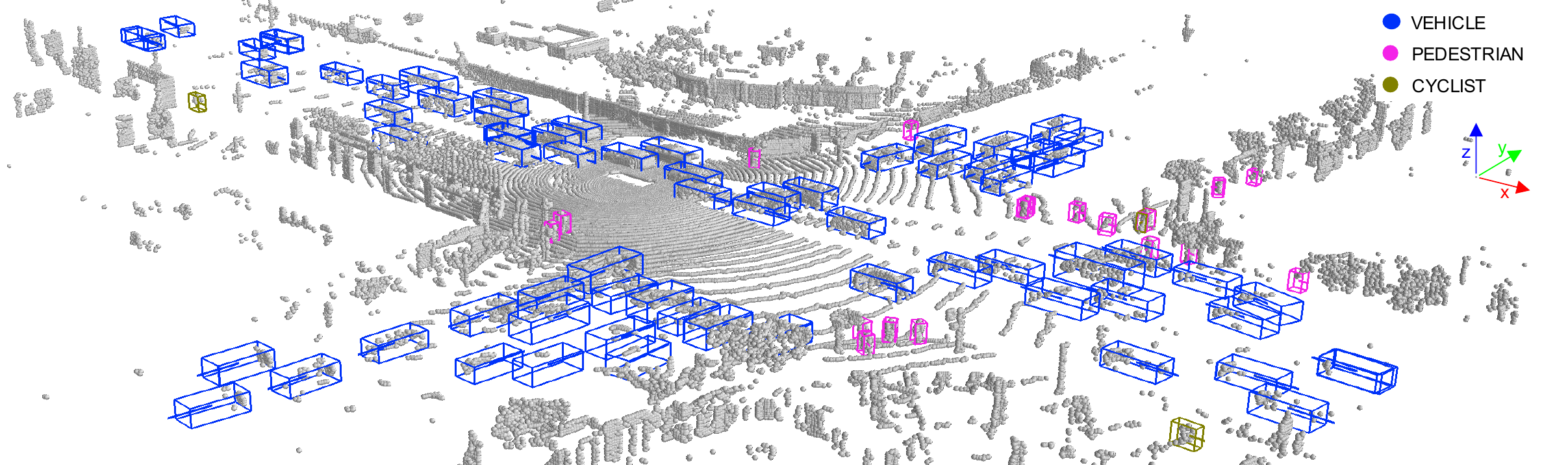}
    \captionof{figure}{Detection results on the testing set of Waymo Open Dataset 3D detection track. Legends indicate the color used for each class.}
    \label{fig:Final result 1}
\end{center}
}] 

\newcommand\blfootnote[1]{%
  \begingroup
  \renewcommand\thefootnote{}\footnote{#1}%
  \addtocounter{footnote}{-1}%
  \endgroup
}

\blfootnote{$\prescript{*}{}{}$These authors contributed equally to this work.}

\begin{abstract}
In this technical report, we introduce our winning solution ``HorizonLiDAR3D'' for the 3D detection track and the domain adaptation track in Waymo Open Dataset Challenge at CVPR 2020. 
Many existing 3D object detectors include prior-based anchor box design to account for different scales and aspect ratios and classes of objects, which limits its capability of generalization to a different dataset or domain and requires post-processing (\eg Non-Maximum Suppression (NMS)). 
We proposed a one-stage, anchor-free and NMS-free 3D point cloud object detector AFDet~\cite{ge2020afdet}, using object key-points to encode the 3D attributes, and to learn an end-to-end point cloud object detection without the need of hand-engineering or learning the anchors. AFDet~\cite{ge2020afdet} serves as a strong baseline in our winning solution and significant improvements are made over this baseline during the challenges. 
Specifically, we design stronger networks and enhance the point cloud data using densification and point painting. To leverage camera information, we append/paint additional attributes to each point by projecting them to camera space and gathering image-based perception information. The final detection performance also benefits from model ensemble and Test-Time Augmentation (TTA) in both the 3D detection track and the domain adaptation track. 
Our solution achieves the \nth{1} place with $77.11\%$ mAPH$/$L2 and $69.49\%$ mAPH$/$L2 respectively on the 3D detection track and the domain adaptation track.

\end{abstract}

\section{Introduction to the Challenge}

The Waymo Open Dataset Challenges at CVPR 2020 which starts from March \nth{19} 2020 and ends on May \nth{31} 2020, is the largest and the most exciting competition of the challenging perception tasks in autonomous driving. The Waymo Open Dataset~\cite{sun2019scalability} was recently released with high-quality data collected from both LiDAR and camera sensors in real self-driving scenarios and enables many new exciting researches. In the 3D detection track, the challenge requires the algorithm to detect the objects as a set of 3D bounding boxes. The domain adaptation track is similar to the 3D detection track except that the data was collected from a different location.


\begin{figure*}
\begin{center}

\includegraphics[width=0.7\textwidth]{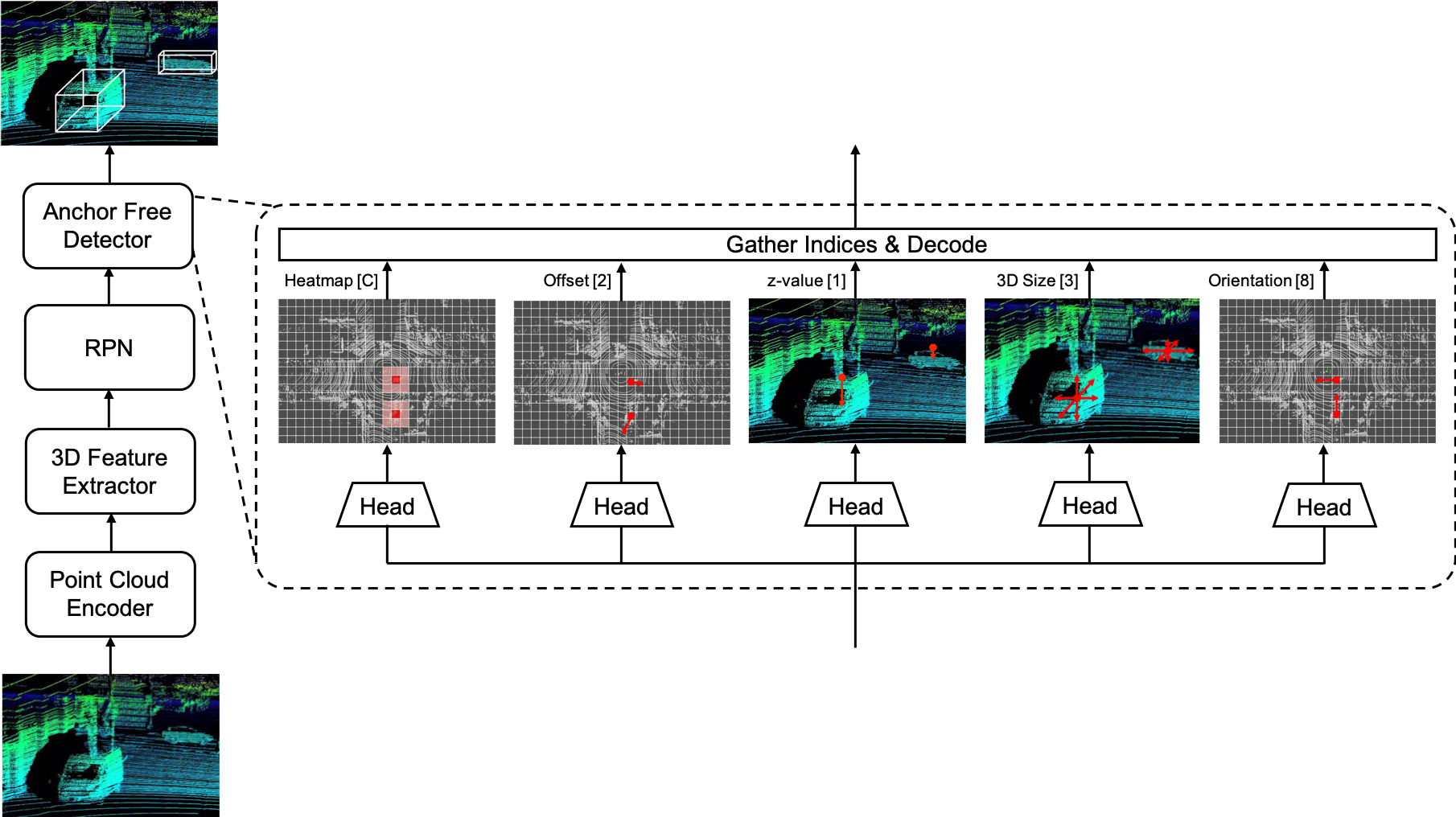}
\end{center}
\caption{The framework of anchor-free one-stage 3D detection (AFDet) system. The whole pipeline consists of the Point Cloud Encoder, 3D Feature Extractor and Region Proposal Network (RPN), and the Anchor-free Detector. The number in the brackets indicates the output channels in the last convolution layer. $C$ is the number of categories used in the detection.}
\label{fig:framework}
\end{figure*}

\section{Network}
In this section, we present the details of our 3D detector in the challenge. 
We use our AFDet~\cite{ge2020afdet}, which is one-stage, anchor-free and NMS-free, as a strong baseline 3D point cloud detector. Significant improvements were made to this baseline during the challenges. Our proposed AFDet consists of four modules: point cloud encoder, 3D Feature Extractor, Region Proposal Network (RPN), and anchor-free detector. We mainly describe the difference between the original AFDet and the one we used in this section.

\subsection{Point Cloud Encoder}

\begin{figure}
\begin{center}

\includegraphics[width=0.50\textwidth]{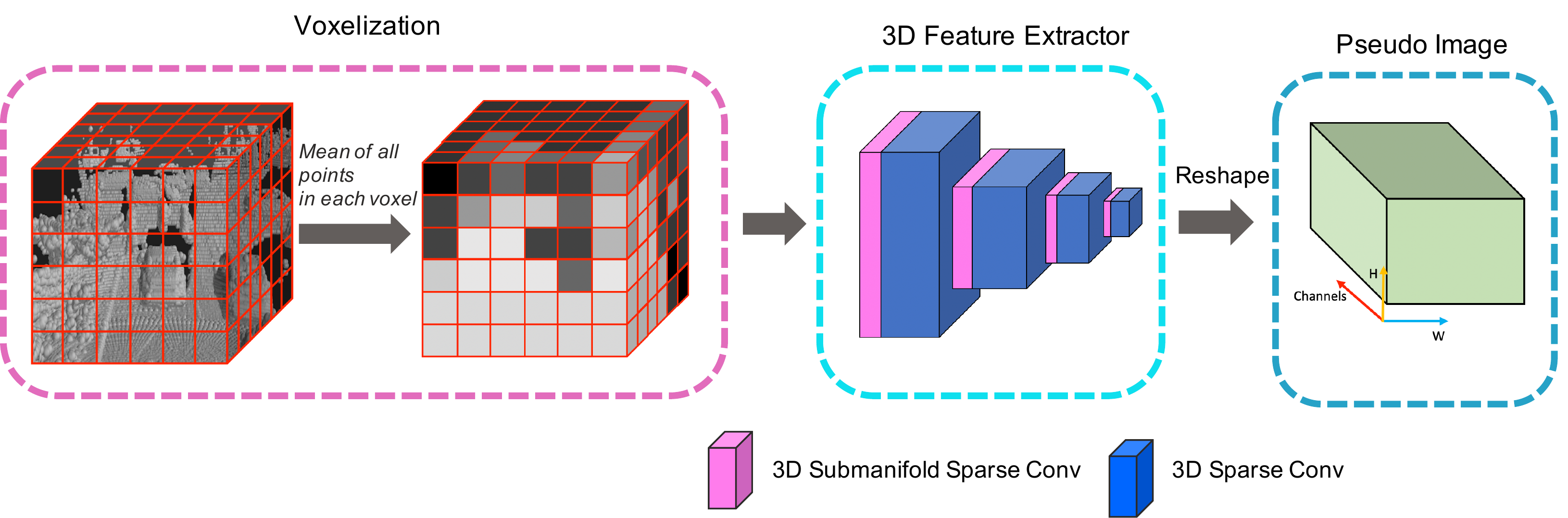}
\end{center}
\caption{Point cloud encoding process from voxelization to pseudo image. 3D Submanifold Sparse Conv and 3D Sparse Conv are used in the 3D Feature Extractor. }
\label{fig:Point cloud encoding}
\end{figure}

We use a simple point cloud encoder~\cite{zhu_det3d} to voxelize the point cloud. Specifically, we use grid size $0.04m$, $0.04m$, $0.1m$ along $x$, $y$, $z$ axis respectively to convert the raw point cloud into voxel presentation. 
In each voxel we calculate the mean of all points inside it and feed the result to 3D feature extractor. The resulting feature map will be reshaped to form a top-down view pseudo image, as illustrated in Figure~\ref{fig:Point cloud encoding}.

\begin{figure*}
\centering
\begin{subfigure}[b]{.3363\linewidth}
\includegraphics[width=\linewidth]{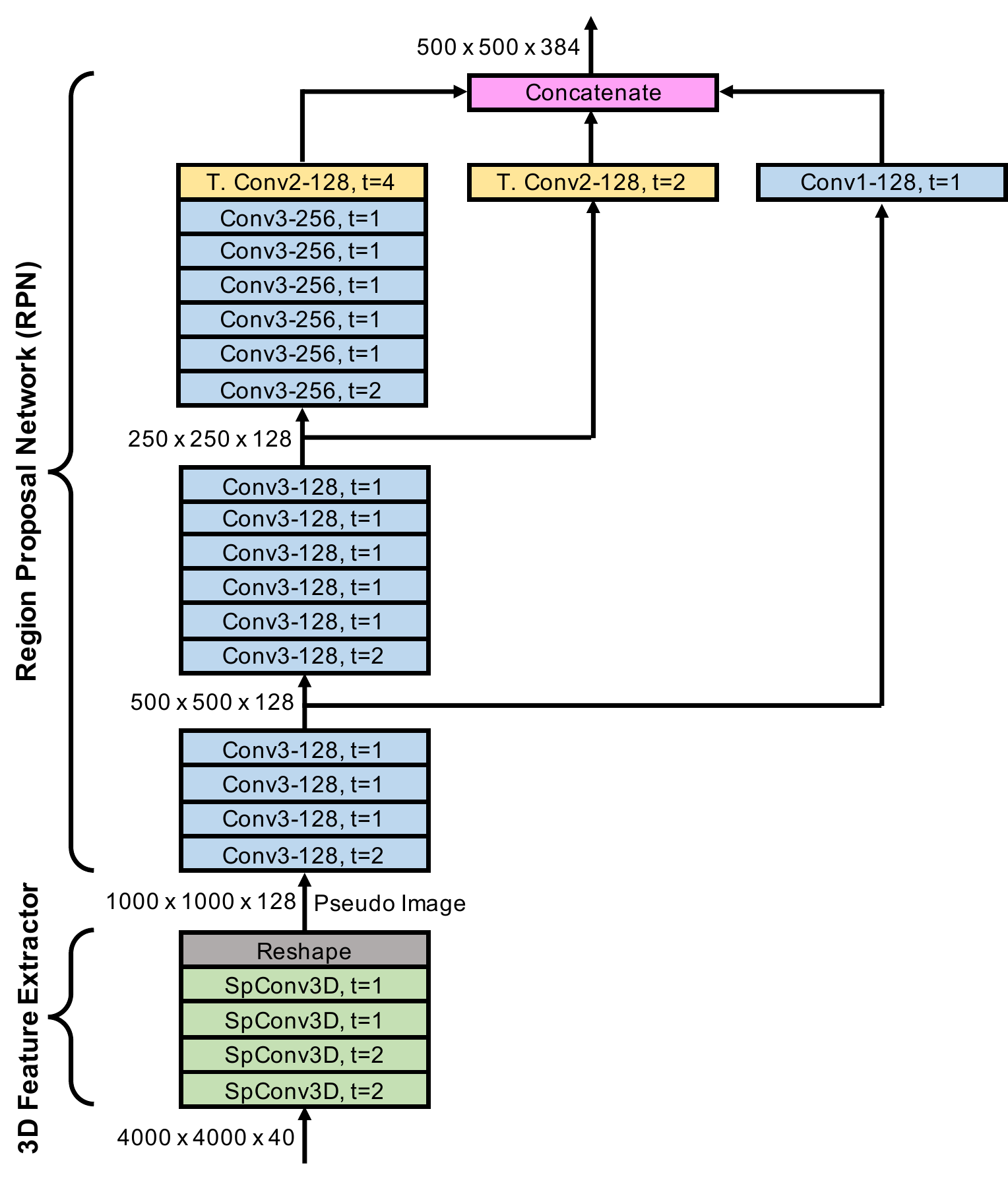}
\caption{B1}\label{fig:B1}
\end{subfigure}
\begin{subfigure}[b]{.3049\linewidth}
\includegraphics[width=\linewidth]{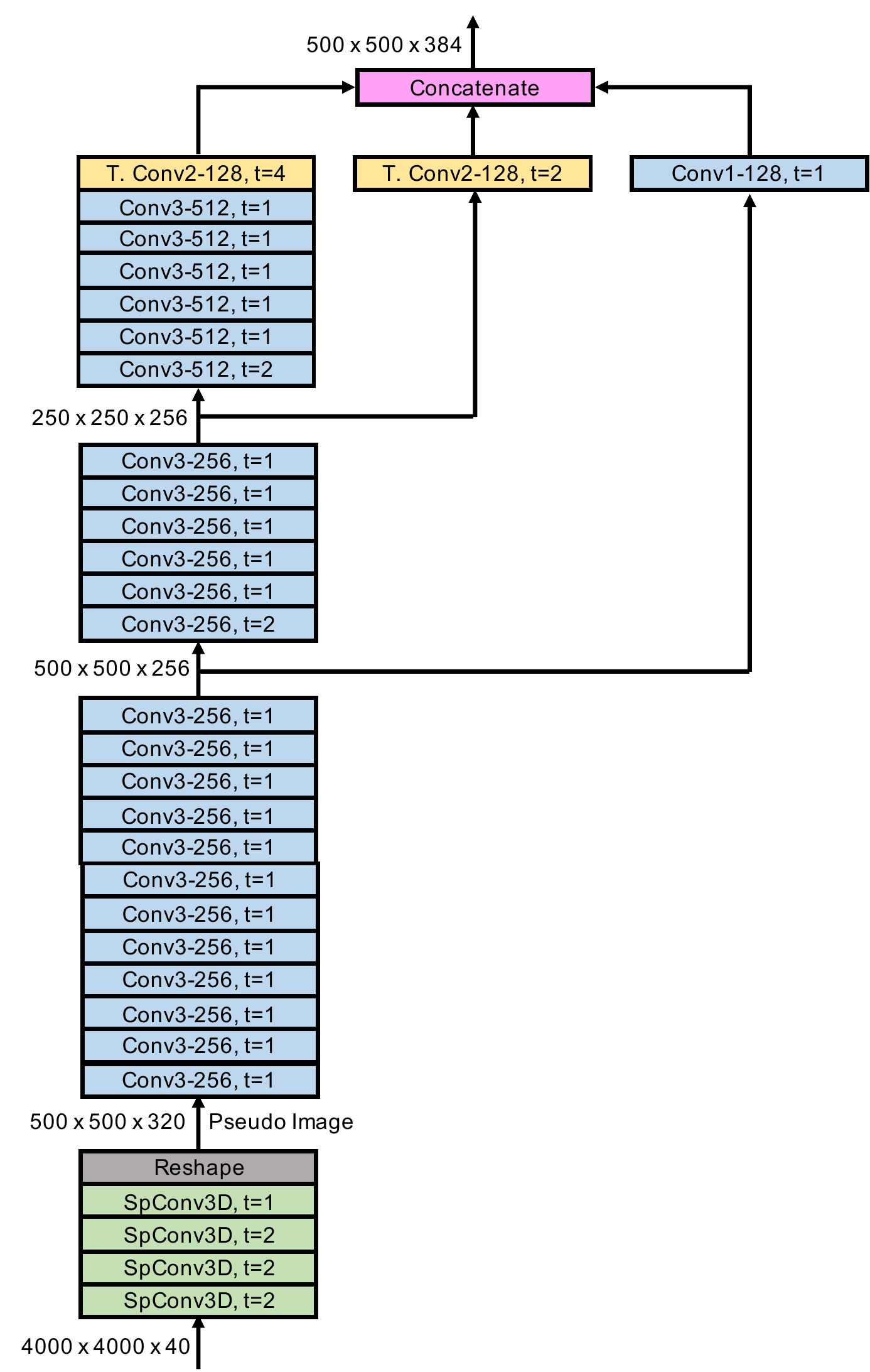}
\caption{B2}\label{fig:B2}
\end{subfigure}
\begin{subfigure}[b]{.3088\linewidth}
\includegraphics[width=\linewidth]{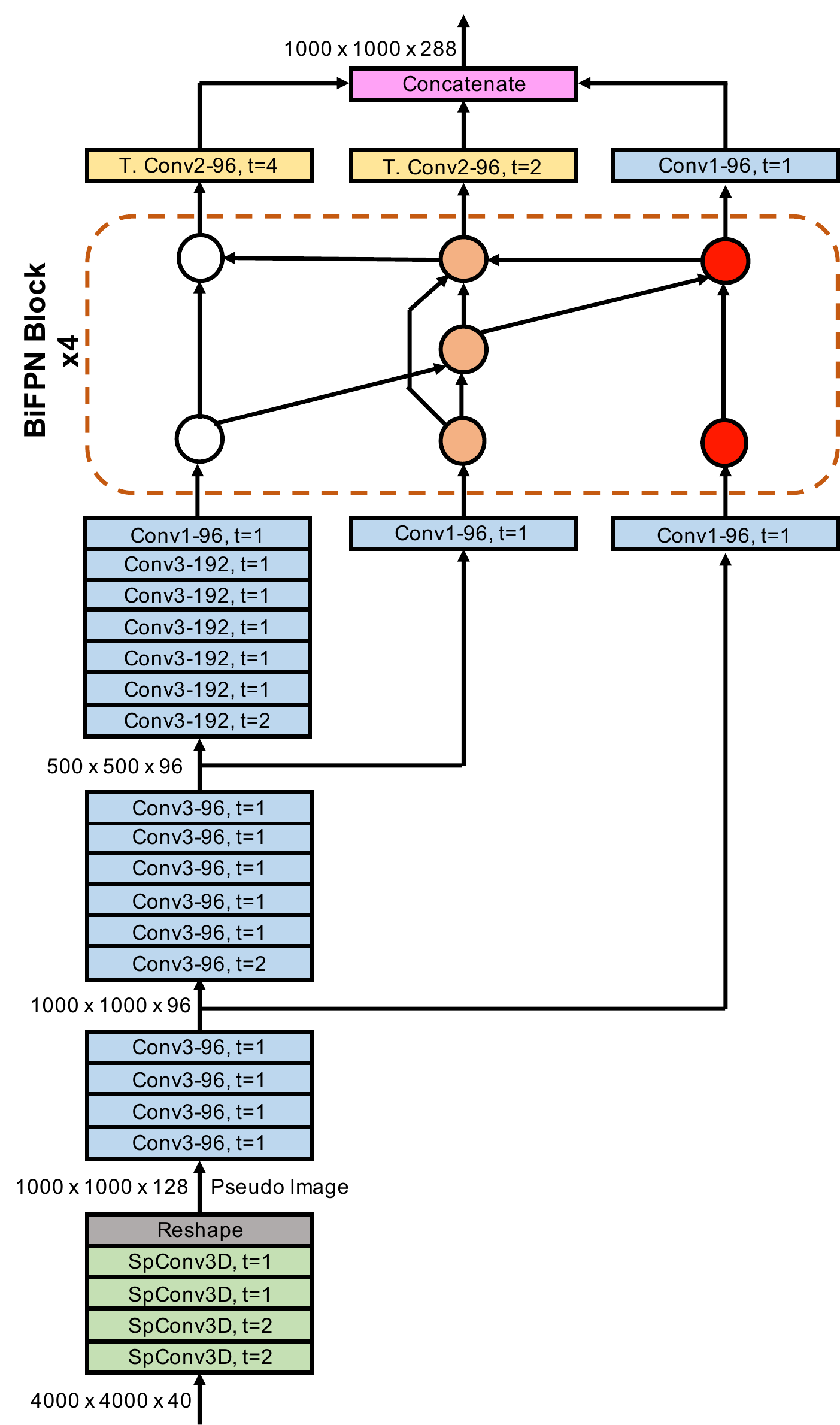}
\caption{B3}\label{fig:B3}
\end{subfigure}
\caption{The illustration and parameters of B1, B2 and B3. Rectangle with different colors represent different operations. Circles represent feature maps. T. Conv is short for transposed convolution. The feature map sizes of each output are their sizes at the inference stage. Better viewed in color and zoomed in for details.}
\label{fig:backbone}
\end{figure*}

\subsection{3D Feature Extractor and RPN}
We introduce two different 3D Feature Extractor \emph{FE-v1}, \emph{FE-v2}, and three different RPN~\cite{yan2018second, zhou2018voxelnet} \emph{RPN-v1}, \emph{RPN-v2}, \emph{RPN-v3}. Their combinations B1, B2 and B3 are used in the final submissions during the challenge.

\emph{SpMiddleFHD} in~\cite{zhu_det3d} are employed as the 3D Feature Extractor in \emph{FE-v1}. 
It contains four phases, and each phase contains several submanifold convolutional layers~\cite{yan2018second} and one sparse convolutional layer~\cite{yan2018second} to perform downsampling along the $z$-axis.
The feature map is downsampled 4$\times$ after \emph{FE-v1}.

For the 3D Feature Extractor in \emph{FE-v2}, we use a similar extractor to~\cite{he2020sassd} with auxiliary branch removed. 
It has four sparse 3D convolution blocks, each of which is composed of sub-manifold convolutions with the kernel size of 3. 
The last three blocks use an additional sparse convolution with a stride of 2 to downsample feature map size. 
Then the feature map is reshaped to obtain the Bird's-Eye View (BEV) representation.
Eight standard 3$\times$3 convolutions are applied to further abstract the feature. The final downsample factor for \emph{FE-v2} is 8.

RPN has several downsample and upsample blocks. In \emph{RPN-v1}, we use the same structure as in~\cite{lang2019pointpillars, yan2018second}.
\emph{RPN-v1} has 3 cascaded downsample blocks, each of which downsamples the feature map by 2. So the feature map of each downsample blocks is downsampled by 2, 4, 8, respectively. Then three upsample blocks are used separately to upsample the feature map by 1, 2, 4. We concatenate them to get the output of the \emph{RPN-v1}. So the final feature map of \emph{RPN-v1} is downsampled 2$\times$.

The \emph{RPN-v2} shares the same structure with \emph{RPN-v1} except for two differences.
The first downsample block uses a downsample factor of 1 instead of 2, and thus the final downsample factor is 1 for \emph{RPN-v2}. 
The other difference is that the channels of each block are doubled compared with \emph{RPN-v1}.

We further improves \emph{RPN-v3} over \emph{RPN-v1}.
Based on \emph{RPN-v1}, the first downsample block uses a factor of 1 instead of 2. Then we reduce the number of filters for each layer to 3/4 (\ie from 128 to 96, from 256 to 192). Inspired by~\cite{tan2019efficientdet}, we add a simplified version of BiFPN with 4 repeating blocks after the downsample blocks in the \emph{RPN-v3}. Unlike the original BiFPN~\cite{tan2019efficientdet} with different number of filters for each repeat, 
we use three separate convolutional layers to produce the same number of filters (96) for the output of each downsample block. Then we repeat the block four times with the same number of filters. Our simple BiFPN does not deploy weighted fusion mechanism~\cite{tan2019efficientdet}. The final downsample factor of \emph{RPN-v3} is 1.

We use three different network, which are B1, B2 and B3, during the challenge. The 3D Feature Extractor and RPN combinations of B1, B2 and B3 can be found in Table~\ref{tbl:backbone}. Their parameters and detailed illustration are shown in Figure~\ref{fig:backbone}. In 3D Feature Extractor and RPN, each convolution is followed by a batch normalization~\cite{pmlr-v37-ioffe15} and ReLU non-linearity.

\begin{table}[t]
\begin{center}
\resizebox{0.46\textwidth}{!}{%
\setlength\tabcolsep{6pt}
\begin{tabular}{l|cc|c}
\hline
                           &3D Feature Extractor                &RPN   & Downsample Factor \\
\hline\hline
B1              &\emph{FE-v1}                           &\emph{RPN-v1} &8    \\
B2                  &\emph{FE-v2}                          &\emph{RPN-v2} &8 \\
B3      &\emph{FE-v1}                            &\emph{RPN-v3}  &4  \\

\hline
\end{tabular}
}
\end{center}
\caption{3D Feature Extractor and RPN combinations of B1, B2 and B3.} 
\label{tbl:backbone}
\end{table}

\subsection{Anchor Free Base Detector}

We briefly introduce the anchor free base detector in the report. For more details, please refer to~\cite{ge2020afdet}. The AFDet~\cite{ge2020afdet} consists of five sub-heads, including the keypoint heatmap head, the local offset head, the $z$-axis location head, the 3D object size head, and the orientation head. Figure~\ref{fig:framework} shows the details of the anchor free detector.

\textbf{Five Sub-Heads.} The heatmap head and the offset head predict a keypoint heatmap $\mathit{\hat{M} \in \mathbb{R}^{W \times H \times C}}$ and a local offset regression map $\hat{O} \in \mathbb{R}^{W \times H \times 2}$ respectively, with $C$ the number of keypoint types. 
The keypoint heatmap helps to locate the object center ($x$-$y$ coordinates) in BEV. 
The offset regression map compensates the discretization error due to voxelization, and helps to recover more accurate object locations in BEV.
The $z$-axis location head regresses the $z$-axis values.
Additionally, we regress the object sizes $\hat{S} \in \mathbb{R}^{W \times H \times 3}$ directly. For orientation prediction, we use the \textit{MultiBin} method following~\cite{Mousavian_2017_CVPR, zhou2019objects}.

\textbf{Loss.} The heatmap head training uses the modified focal loss~\cite{lin2017focal}. For the orientation prediction head, the classification part is trained with softmax while the offset part is trained with $L_1$ loss.  $L_1$ loss is employed in the local offset head, the $z$-axis location head, and the 3D object size head training. The overall training objective is
\begin{equation}
\mathit{\mathcal{L} = \mathcal{L}_{heat} + \lambda_{off}\mathcal{L}_{off} + \lambda_{z}\mathcal{L}_{z} + \lambda_{size}\mathcal{L}_{size} + \lambda_{ori}\mathcal{L}_{ori}},
\end{equation}
where $\lambda$ represents the weight for each sub-task. For all regression sub-tasks including local offset, $z$-axis location, size and orientation, we only regress $N$ objects which are in the detection range.

\textbf{Gather Indices and Decode.} 
From the resultant keypoint heatmap, we can easily decode the center $\left ( \hat{x}, \hat{y}\right)$ of each object, along with the local offset correcting the discretization error. Other information such as orientation and dimension can be obtained from the regression results~\cite{ge2020afdet}.

\section{Methods}
The network architecture has been explained in details in the previous section. In this section, we first introduce our input format and data augmentation strategies. Then we explain a novel painting method using 2D detection, test time augmentation, and model ensemble.
\subsection{Input and Data Augmentation}
As in ~\cite{caesar2019nuscenes}, we accumulate LiDAR sweeps to utilize temporal information and to densify the LiDAR point cloud. To distinguish points from different sweeps, time difference $\mathit{\Delta t}$ is attached to the point cloud  as an additional attribute. 
In Waymo challenge, we use past four frames combined with the current frame as our input point cloud. A Detailed ablation study can be found in Table~\ref{tbl:performance1}.

Waymo Open Dataset~\cite{sun2019scalability} provides five kinds of LiDAR sweeps. To fully utilize the information, we combine all the first and second returns point cloud generated by all five LiDARs. Specifically, considering frame densification and painting, our input format should be $\mathit{(x,y,z, reflectance, [...painted\_scores...], \Delta t)}$, with $\mathit{\Delta t}$ being the time difference as mentioned above.

We use data augmentation strategy following~\cite{yan2018second, lang2019pointpillars}. First, we generate an annotation database containing labels and associated point cloud data. During training, we randomly select 6, 8 and 10 ground truth samples for vehicle, pedestrian and cyclist respectively, and place them into the current frame. Second, we do randomly flipping along $z$-axis~\cite{yang2018pixor}, global rotation following $\mathcal{U}\left ( -\frac{\pi}{4}, \frac{\pi}{4} \right )$, global scaling following $\mathcal{U}\left ( 0.95, 1.05 \right )$ and global translation along $x,y,z$-axis following $\mathcal{U}\left ( -0.2m, 0.2m \right )$~\cite{zhou2018voxelnet, yan2018second, lang2019pointpillars}.
\begin{figure}
\begin{center}

\includegraphics[width=0.50\textwidth]{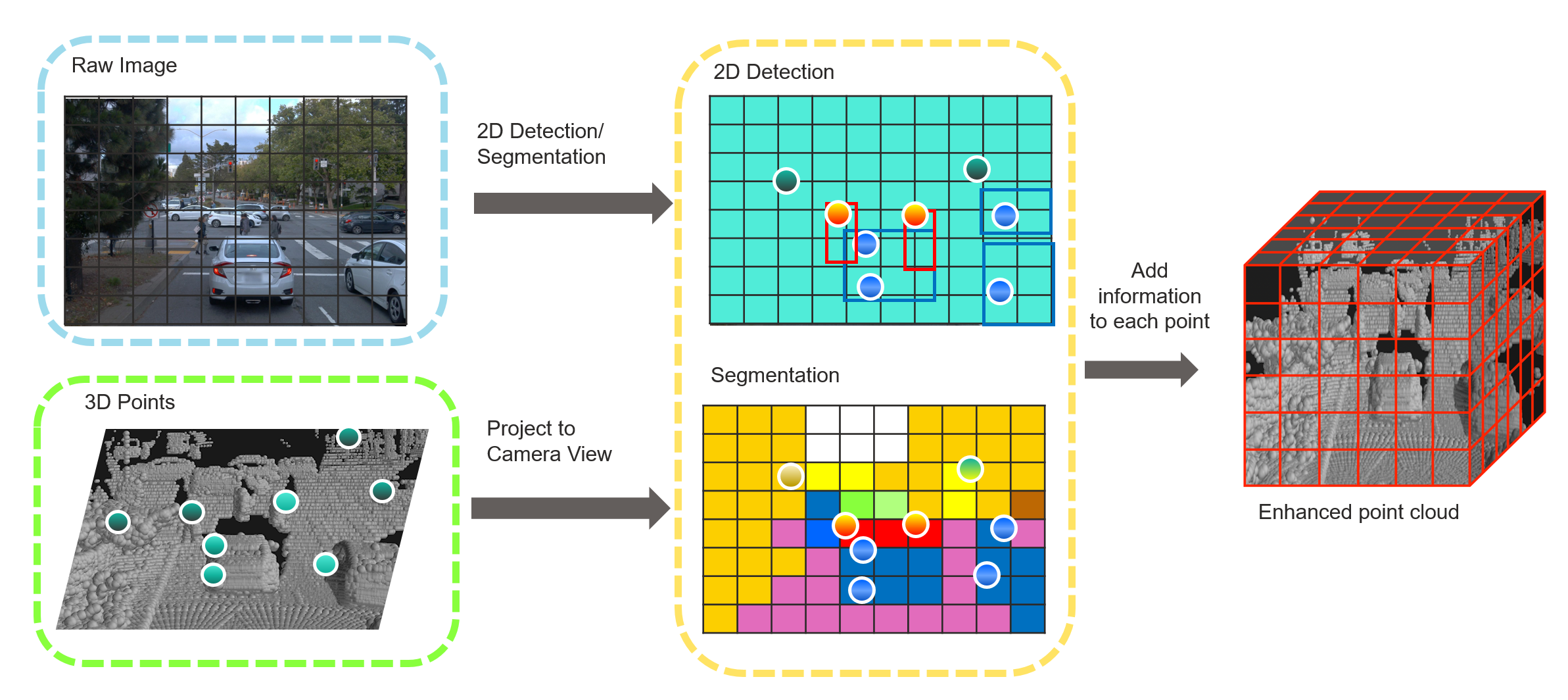}
\end{center}
\caption{Point cloud painting using 2D detections and semantic segmentation labels}
\label{fig:Box and Seg painting}
\end{figure}

\subsection{Painting}
Inspired by PointPainting~\cite{vora2019pointpainting}, we leverage camera information by projecting each point to the image space and gathering its image-based perception information (\ie 2D detections and semantic segmentation labels) which will be appended/painted to the point, as illustrated in Figure~\ref{fig:Box and Seg painting}.

First we use 2D bounding boxes to paint the points. Specifically, points are projected from the vehicle coordinate to the corresponding image frame. If a certain point falls inside a 2D bounding box, the classification information will be appended to the point as extra attributes. Otherwise, zeros will be filled in additional channels. 
The training set is painted with ground truth 2D bounding boxes, while during inference painting uses 2D detections produced by a single Cascade R-CNN based detector used in our 2D detection track. 

Second, we use semantic segmentation information to paint point clouds. We start with a WideResNet38 pretrained on ImageNet~\cite{krizhevsky2012imagenet} and finetune the network on KITTI~\cite{Geiger2012CVPR}. Only around half of the points can be painted in a frame because Waymo Open Dataset only provides images from front and side cameras. The improvements of introducing point painting can be seen in Table~\ref{tbl:performance1}. 


\begin{table}[!h]
\begin{center}
\vspace{5pt}
\setlength\tabcolsep{4pt}
\begin{tabular}{l|c|c c|c c}
\hline
 \multirow{2}{*}{Models} &\multirow{2}{*}{Frames} &\multicolumn{2}{|c|}{Painting} &\multicolumn{2}{|c}{L2}\\
  & &2D Box & Seg &mAP &mAPH \\
\hline
\hline
AFDet-B1 &$\mathcal[-0, +0]  $ &\xmark  &\xmark  &60.87 &56.50   \\
AFDet-B1 &$\mathcal[-4, +0]  $ &\xmark  &\xmark  &65.71 &62.22   \\
AFDet-B1 &$\mathcal[-4, +0]  $ &\cmark  &\xmark  &67.88 &64.77   \\
AFDet-B1 &$\mathcal[-4, +0]  $ &\cmark  &\cmark &68.03 &65.02    \\

\hline
\end{tabular}
\end{center}
\caption{3D detection results of single model on validation set of the 3D detection track. Models are trained on 1/20 of training set and tested on 1/10 of validation set.} 
\label{tbl:performance1}
\end{table}

\begin{figure}
\begin{center}

\includegraphics[width=0.48\textwidth]{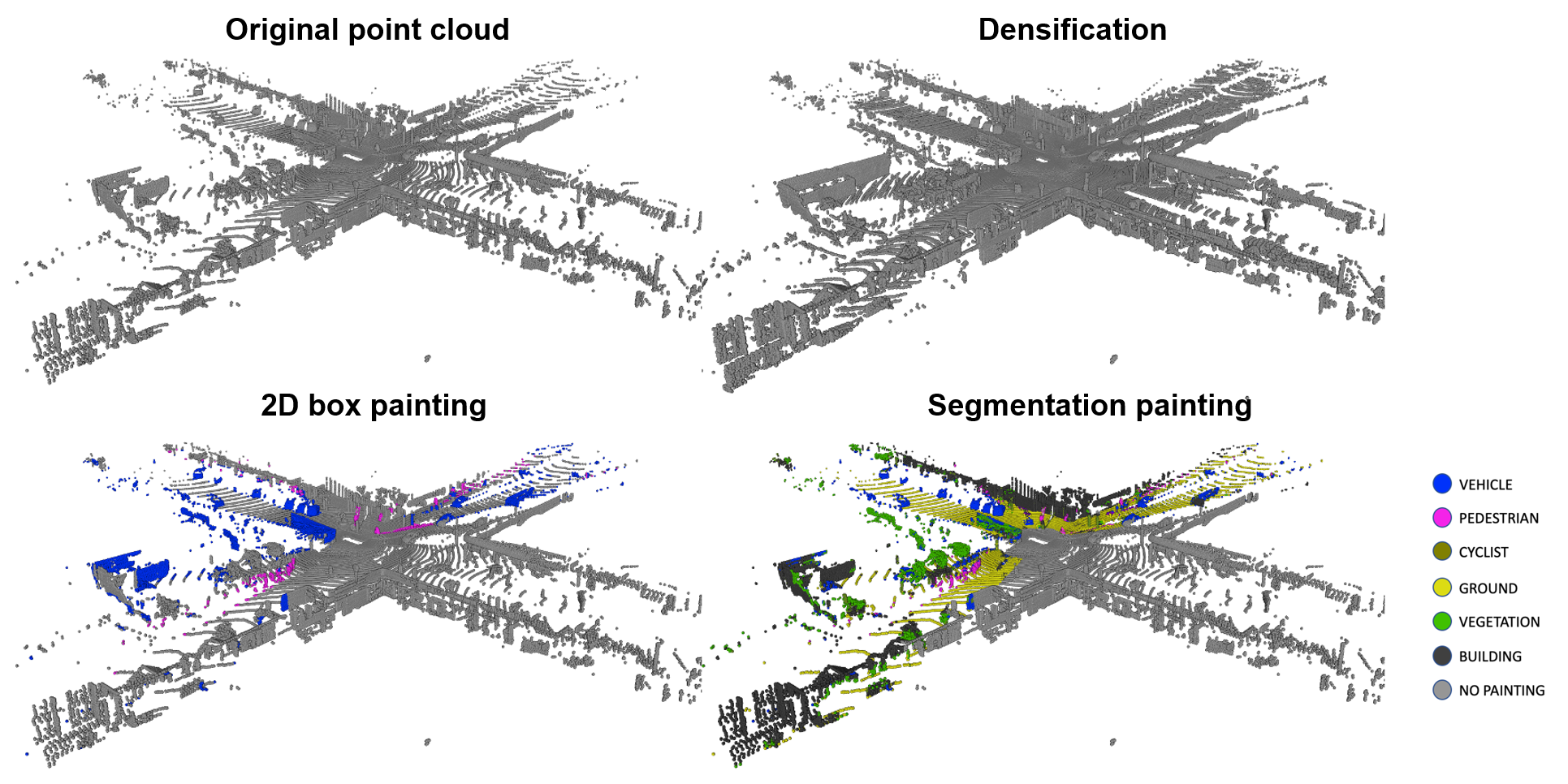}
\end{center}
\caption{Point cloud enhancement. The top left image shows the raw point cloud. The top right image shows point cloud densified with [-4, +0] frames. The bottom left image shows point cloud painted with 2D detection boxes. The bottom right image shows point cloud painted with semantic segmentation. Note that only half of the points are painted in two bottom images because rear-view images are unavailable.}
\label{fig:Input data with different augmentation}
\end{figure}


\subsection{Further Improvements}

\textbf{Test Time Augmentation (TTA).} We perform several different test-time augmentations, including point cloud rotation around pitch, roll and yaw axis, point cloud global scaling and point cloud translation along $z$-axis, which is similar to the data augmentation in the training process. We adapted the Weighted Boxes Fusion (WBF) \cite{solovyev2019weighted} to a 3D variant to merge different groups of detections into the final result. We use $\pm 0.5\degree$ for both pitch and roll rotation, [$0\degree$, $\pm 22.5\degree$, $\pm 45\degree$,  ,$\pm 135\degree$,$\pm 157.5\degree$, $180\degree$] for yaw rotation, [0.95, 1.05] for global scaling, and $\pm 0.2m$ for translation along $z$-axis. 
Our experiments show that the rotation yaw is the most effective one among all the augmentation methods. Therefore we use it in combination with the other augmentation methods. 

\textbf{Ensemble Method and Naive Grid Search.}
As mentioned above, we use a 3D version of WBF \cite{solovyev2019weighted} to ensemble different models with test time augmentation. Equal weights are assigned to AFDet-B1, AFDet-B2 and AFDet-B3. We replace 3D IoU with BEV IoU to speed up the ensembling process. A naive grid search is conducted to select the best thresholds used in the ensembling. We search three variables including $\theta_{IoU}$ (box associating IoU threshold), $\theta_{s1}$ (pre-wbf box skipping threshold) and $\theta_{s2}$ (post-wbf box skipping threshold). Search spaces for $\theta_{IoU}$, $\theta_{s1}$ and $\theta_{s2}$ are $\mathcal[0.40, 0.80]  $, $\mathcal[0.00, 0.25]  $, $\mathcal[0.01, 0.20]  $ respectively. Search intervals for $\theta_{IoU}$, $\theta_{s1}$ and $\theta_{s2}$ are 0.05, 0.05 and 0.01 respectively. $\theta_{s2}$ is searched from 0.01 instead of 0.00 to accelerate the search process. All grid searches are performed on 1/10 of the validation set. We use different thresholds for different classes as shown in Table~\ref{tbl:thresh selection}.

\begin{table}[t]
\begin{center}
\setlength\tabcolsep{3pt}
\begin{tabular}{l|c|c|c|c}
\hline
  Dataset     &  Thres.   &Vehicle  &Pedestrian  &Cyclist\\
\hline\hline
 \multirow{3}{*}{3D Detection} &\multicolumn{1}{|c|}{$\theta_{IoU}$ } &\multicolumn{1}{|c|}{0.80} &\multicolumn{1}{|c|}{0.70} &\multicolumn{1}{|c}{0.65}\\
  &$\theta_{s1}$ &0.10 & 0.15 &0.25 \\
    &$\theta_{s2}$    &0.03   &0.03  &0.03        \\

\hline
 \multirow{3}{*}{Domain Adaptation} &\multicolumn{1}{|c|}{$\theta_{IoU}$ } &\multicolumn{1}{|c|}{0.80} &\multicolumn{1}{|c|}{0.70} &\multicolumn{1}{|c}{0.65}\\
  &$\theta_{s1}$ &0.10 & 0.15 &0.25 \\
    &$\theta_{s2}$    &0.05   &0.05  &0.05        \\

\hline
\end{tabular}
\end{center}
\caption{Thresholds used in the final submission. $\theta_{IoU}$, $\theta_{s1}$ and $\theta_{s2}$ are box associating IoU threshold, pre-wbf box skipping threshold and post-wbf box skipping threshold.} 
\label{tbl:thresh selection}
\end{table}

\begin{table}[!h]
\begin{center}

\vspace{5pt}
\setlength\tabcolsep{2pt}
\begin{tabular}{l|c|c|c c}
\hline
 \multirow{2}{*}{Models} &\multicolumn{1}{|c|}{Test Time} &\multicolumn{1}{|c|}{Naive} &\multicolumn{2}{|c}{L2}\\
  &Augmentation &Grid Search &mAP &mAPH\\
\hline
\hline
AFDet-B1 &\xmark &\xmark &68.03 &65.02 \\
AFDet-B2 &\xmark &\xmark &67.05 &64.29 \\
AFDet-B3 &\xmark &\xmark &68.91 &65.24 \\
B1 + B2 &\xmark &\xmark &67.97 & 65.71\\
B1 + B3 &\xmark &\xmark &69.72 & 67.16\\
B1 + B2 + B3 &\xmark &\xmark &71.08 &68.96 \\
B1 + B2 + B3 &\cmark &\xmark &73.65 &72.14 \\
B1 + B2 + B3 &\cmark &\cmark &75.21 & 73.63 \\

\hline
\end{tabular}
\end{center}
\caption{3D detection ensemble results on validation set of the 3D detection track. Models are trained on 1/20 of training set and tested on 1/10 of validation set.} 
\label{tbl:performance2}
\end{table}

\begin{table*}[h]
\begin{center}
\resizebox{0.9\textwidth}{!}{%
\vspace{5pt}
\setlength\tabcolsep{4pt}
\begin{tabular}{l|c c|c c|c c|c c}
\hline
 \multirow{2}{*}{Models}  &\multicolumn{2}{|c|}{VEHICLE} &\multicolumn{2}{|c}{PEDESTRIAN} &\multicolumn{2}{|c}{CYCLIST} &\multicolumn{2}{|c}{ALL}\\
  &L2 mAP &L2 mAPH &L2 mAP &L2 mAPH &L2 mAP &L2 mAPH &L2 mAP &L2 mAPH\\
\hline
\hline
HorizonLiDAR3D (Ours) &\textbf{78.23}  &\textbf{77.83} &\textbf{79.32} &\textbf{76.50} &\textbf{77.91}  &\textbf{76.98} &\textbf{78.49} &\textbf{77.11}  \\
PV-RCNN  &73.69  &72.23 &73.98 &70.16 &72.38  &71.16 &73.35 &71.52  \\
TS-LidarDet &72.65  &72.12 &68.10 &59.32 &66.55  &65.16 &69.10 &65.53   \\
Simple Baseline v2 &66.44  &65.91 &66.00 &60.93 &65.13  &64.10 &65.84 &63.65  \\
Det3D-Waymo-3D-FS-VS  &66.03  &65.11 &66.38 &57.83 &67.60  &66.18 &66.67 &63.04   \\
\hline
\end{tabular}
}
\end{center}
\caption{Top five submissions of Waymo Open Dataset Challenge on 3D detection track.} 
\label{tbl:3d detection test result}
\end{table*}


\begin{table*}[h]
\begin{center}
\resizebox{1\textwidth}{!}{%
\vspace{5pt}
\setlength\tabcolsep{2pt}
\begin{tabular}{l|c c|c c|c c|c c|c c}
\hline
 \multirow{2}{*}{Models}  &\multicolumn{2}{|c|}{VEHICLE} &\multicolumn{2}{|c}{PEDESTRIAN} &\multicolumn{2}{|c}{CYCLIST} &\multicolumn{2}{|c}{ALL}
 &\multicolumn{2}{|c}{DOMAIN GAP (ALL)}\\
  &L2 mAP &L2 mAPH &L2 mAP &L2 mAPH &L2 mAP &L2 mAPH &L2 mAP &L2 mAPH &L2 mAP &L2 mAPH\\
\hline
\hline
HorizonLiDAR3D (Ours) &\textbf{60.39}  &\textbf{60.04} &\textbf{64.00} &\textbf{61.63} &\textbf{87.55}  &\textbf{86.79} &\textbf{70.65} &\textbf{69.49} &\textbf{7.84} & \textbf{7.62}  \\
PV-RCNN-DA  &59.67  &59.08 &48.27 &45.74 &28.31  &27.50 &45.42 &44.11& 27.93& 27.41  \\
Simple Baseline v2  &48.47  &47.92 &46.13 &43.21 &18.84  &18.71 &37.81 &36.62&28.03 &27.02  \\
Det3D-Waymo-DA  &50.07  &49.33 &43.51 &37.90 &4.75  &4.66 &32.78 &30.63&33.89 & 32.41  \\
Simple Baseline  &45.50  &44.77 &45.99 &43.12 &1.53  &1.48 &31.01 &29.79  &- &- \\
\hline
\end{tabular}
}
\end{center}
\caption{Top five submissions of the domain adaptation track. Domain gap is the difference between 3D detection track and domain adaptation track.} 
\label{tbl:domain adaptaion test result}
\end{table*}


\section{Experiment Settings}

We implemented our detector based on the Det3D framework~\cite{zhu_det3d}. To reduce the GPU memory usage, we train our detector within range [ ( -76.8, 76.8), ( -51.2, 51.2), ( -1, 3)] respect to $x,y,z$-axis, while enlarging the detection range to [ ( -80, 80), ( -80, 80), ( -1, 3)] at inference. The max number of objects is set to $300$. For all of our models, we set max point per voxel to $5$, max voxel num to $1,000,000$. For the offset regression head, we use $r=2$ as default to regress a square area with side length $5$.
We use max pooling with the kernel size $3 \times 3$, stride 1, and apply AND operation between the feature map before and after the max pooling to get the peaks of the keypoint heatmaps at the inference stage. 
Therefore there is no need to suppress overlapped detections. The weight we use for different sub-losses are $\lambda_{off}=1.0$, $\lambda_{z}=1.5$, $\lambda_{size}=0.3$ and $\lambda_{ori}=1.0$. All parameters we list here are their default values unless explicitly stated otherwise.

We use AdamW~\cite{loshchilov2018decoupled} optimizer with one-cycle policy~\cite{one_cycle}. We set learning rate max to $3 \times 10^{-3}$, division factor to 2, momentum ranges from 0.95 to 0.85, fixed weight decay to 0.01 to achieve convergence. To quickly verify our idea, we sample the training set every 20 frames, and validation set every 10 frames according to their timestamps. Unless we explicitly indicate, all models used for ablation study, including Table~\ref{tbl:performance1} and Table~\ref{tbl:performance2}, are trained on 1/20 training data, tested on 1/10 validation data. In the final submission, we first train 60 epochs on 1/20 training data and finetune 10 epochs on the whole trainval data for AFDet-B1 and AFDet-B2. We only train 60 epochs on 1/20 training data and finetune 3 epochs on the whole trainval data for AFDet-B3. Besides, we adopt half-precision (FP16)~\cite{micikevicius2017mixed, amp} for B3 model to reduce GPU memory usage and speed up the training process. We ensemble AFDet-B1, AFDet-B2 and AFDet-B3 for 3D detection and AFDet-B1 and AFDet-B2 for domain adaptation in our final submission.

\begin{table}[!h]
\begin{center}
\resizebox{0.48\textwidth}{!}{%
\vspace{5pt}
\setlength\tabcolsep{4pt}
\begin{tabular}{l|c c|c c || c c|c c}
\hline
 \multirow{2}{*}{} &\multicolumn{2}{|c|}{3D training} &\multicolumn{2}{|c||}{3D validation} &\multicolumn{2}{|c|}{Domain training} &\multicolumn{2}{|c}{Domain validation}\\
  &Num &Per &Num &Per &Num &Per  &Num &Per \\
\hline
\hline
Object               &6.3M  &100\%    &1.6M  &100\%      &260K   &100\%      &45.2K    &100\%   \\
Vehicle              &4.2M  &67.15\%  &1.1M  &68.42\%    &242K   &93.15\%    &43K      &95.11\% \\
Pedestrian           &2.0M  &32.07\%  &0.5M  &30.80\%    &17.6K  &6.78\%     &2.2K     &4.89\%  \\
Cyclist              &49K &0.78\%   &12.3K &0.78\%     &184    &0.07\%     &0        &0       \\

\hline
\end{tabular}
}
\end{center}
\caption{Number of objects and the percentage of each class in the 3D detection and domain adaptation tracks.}  
\label{tbl:stat}
\end{table}
\begin{figure*}
\begin{center}

\includegraphics[width=1.0\textwidth]{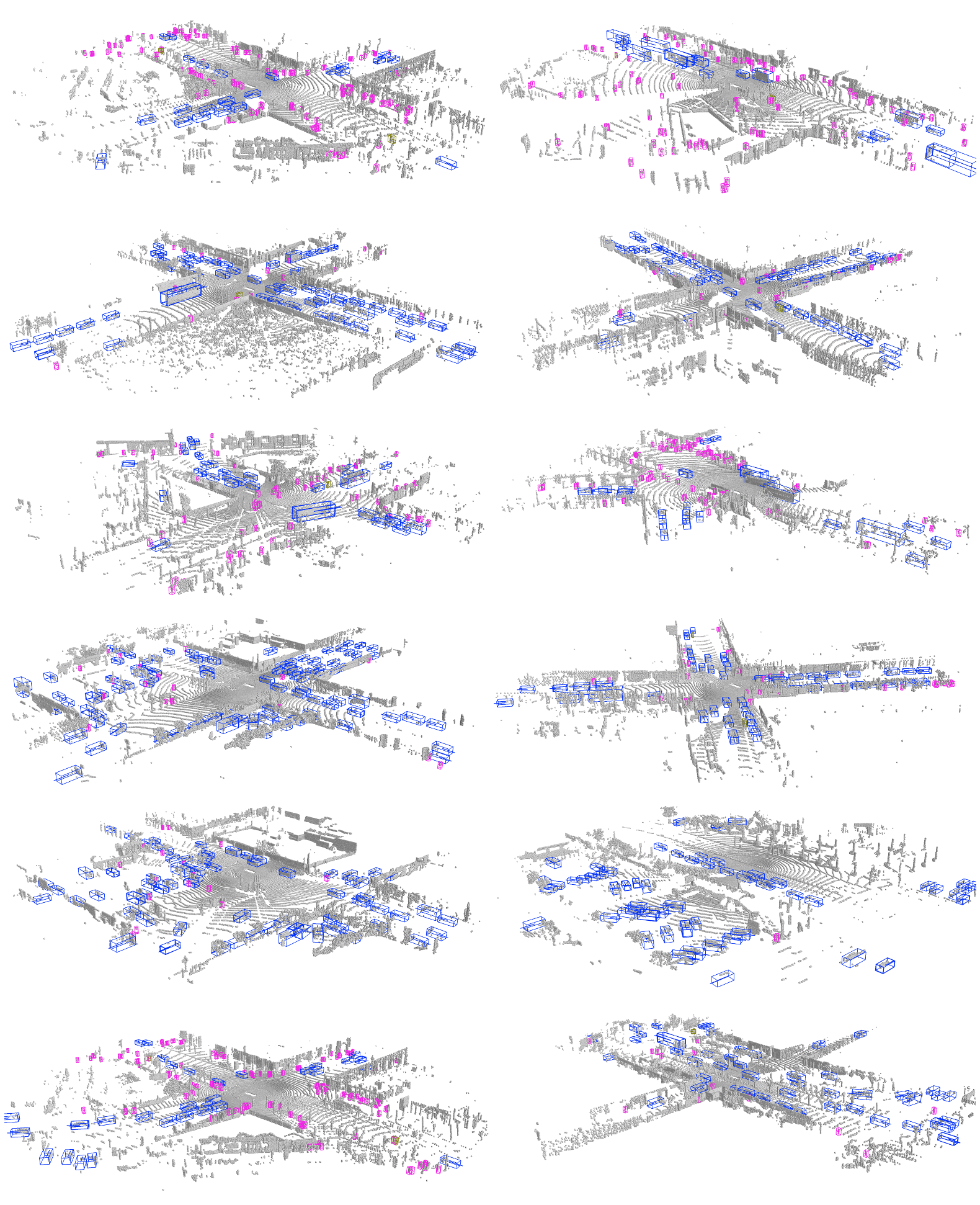}
\end{center}
\caption{Examples of results on testing set of the 3D detection track, only bounding boxes with score larger than 0.15 are visualized.}
\label{fig:Final result 1}
\end{figure*}

\section{Results}
\textbf{3D Detection.}
To study the effect of each module used in our solution, we conduct ablation experiments on the 1/10 validation set as shown in Table~\ref{tbl:performance1} and Table~\ref{tbl:performance2}. Specifically, Table~\ref{tbl:performance1} demonstrates the contributions made by densification and point painting, while Table~\ref{tbl:performance2} shows the improvements introduced by ensembling the networks and using TTA and WBF. The final results on the official 3D detection leaderboard are shown in Table~\ref{tbl:3d detection test result}. As can be seen from Table~\ref{tbl:3d detection test result}, our method outperformed the \nth{2} place solution by $5.59\%$ and the \nth{3} place solution by $11.58\%$ in terms of mAPH L2. 

\textbf{Domain Adaptation.}
Waymo Open Dataset~\cite{sun2019scalability} also provides additional data collected from different locations with only a small subset annotated to test the robustness against domain diversity. We directly apply our models trained for the 3D track to the domain adaptation data. For the final submission, AFDet-B1 and AFDet-B2 were applied with  test time augmentation and the results were merged by 
WBF. $\theta_{IoU}$, $\theta_{s1}$ and $\theta_{s2}$ for our submission are listed in Table~\ref{tbl:thresh selection}. The results show that our solution outperforms other methods in domain adaptation by a big margin as shown in Table~\ref{tbl:domain adaptaion test result}.

This result demonstrated the benefits of using anchor-free model and the better generalizability of our method and its robustness against domain diversity. Another reason might be due to the fact that most other methods finetuned their models using annotated domain adaptation data, which is highly imbalanced. As shown in Table~\ref{tbl:stat}, we compared the statistics computed from 3D detection and domain adaptation datasets respectively. For example, only 0.07\% of the objects are cyclists in the domain adaptation training set, which is 10$\times$ less than the 0.78\% in the 3D detection training set. Fine-tuning on the domain adaptation dataset might compromise the performance of the resulting detector, especially for rare classes such as cyclist. Leveraging image information may also contributed to the robustness of our solution against domain diversity. 

\section{Conclusion}
A state-of-the-art 3D detection framework is proposed and achieved the \nth{1} place on both 3D detection track and domain adaptation track in the Waymo Open Dataset Challenges at CVPR 2020.

{\small
\bibliographystyle{ieee_fullname}
\bibliography{egbib}
}

\end{document}